# HeartAgent: An Autonomous Agent System for Explainable Differential Diagnosis in Cardiology

Shuang Zhou[1], Kai Yu[1], Song Wang[2], Wenya Xie[3], Zaifu Zhan[4], Meng-Han Tsai[5], Yuen-Hei Chung[6], Shutong Hou[7], Huixue Zhou[1], Min Zeng[1], Bhavadharini Ramu[8], Lin Yee Chen[9], Feng Xie[1], Rui Zhang[1,*]

**Affiliations:**
1. Division of Computational Health Sciences, Department of Surgery, University of Minnesota, Minneapolis, MN, USA
2. Department of Computer Science, University of Central Florida, Orlando, FL, USA
3. College of Science and Engineering, University of Minnesota, Minneapolis, MN, USA
4. Department of Electrical and Computer Engineering, University of Minnesota, Minneapolis, MN, USA
5. Division of Cardiothoracic Surgery, Department of Surgery, University of Colorado Anschutz Medical Campus, Aurora, CO, USA
6. Division of Cardiac Electrophysiology, University of California San Francisco, San Francisco, CA, USA
7. School of Dentistry, University of Minnesota, Minneapolis, MN, USA
8. Division of Cardiology, Department of Medicine, University of Minnesota, Minneapolis, MN, USA
9. Lillehei Heart Institute and Department of Medicine, University of Minnesota, Minneapolis, MN, USA

# Abstract

Heart diseases remain a leading cause of morbidity and mortality worldwide, necessitating accurate and trustworthy differential diagnosis. However, existing artificial intelligence-based diagnostic methods are often limited by insufficient cardiology knowledge, inadequate support for complex reasoning, and poor interpretability. Here we present HeartAgent, a cardiology-specific agent system designed to support a reliable and explainable differential diagnosis. HeartAgent integrates customized tools and curated data resources and orchestrates multiple specialized sub-agents to perform complex reasoning while generating transparent reasoning trajectories and verifiable supporting references. Evaluated on the MIMIC dataset and a private electronic health records cohort, HeartAgent achieved over 36% and 20% improvements over established comparative methods, in top-3 diagnostic accuracy, respectively. Additionally, clinicians assisted by HeartAgent demonstrated gains of 26.9% in diagnostic accuracy and 22.7% in explanatory quality compared with unaided experts. These results demonstrate that HeartAgent provides reliable, explainable, and clinically actionable decision support for cardiovascular care.

# Introduction

Heart diseases, defined as conditions that impair the structural integrity or functional performance of the heart, constitute one of the most prevalent and clinically significant disease categories in modern medicine. In the United States, an estimated 20–25 million adults live with some form of heart disease [1], which remains the leading cause of death, accounting for over 600,000 deaths annually [2,3]. Beyond high mortality, heart diseases are associated with substantial chronic morbidity, extensive healthcare expenditures exceeding $400 billion per year [4], and significant reductions in quality of life, underscoring the critical need for improved diagnostic accuracy in clinical practice [5].

Differential diagnosis in cardiology refers to the process of identifying and prioritizing multiple plausible disease etiologies that could explain a patient's presenting symptoms and clinical findings [6,7]. This process requires integrating and interpreting heterogeneous clinical data, including clinical notes, electrocardiogram (ECG), and relevant cardiac imaging, as well as complex clinical reasoning [8,9], since both cardiac and non-cardiac conditions often present with overlapping characteristics [10]. For instance, a patient presenting with acute chest pain, dyspnea, and profuse diaphoresis may be suspected of having aortic dissection, pulmonary embolism, or myocardial infarction [11]. Despite similar initial presentations, these conditions differ substantially in etiology, disease severity, therapeutic strategies, and prognosis [11,12]. Therefore, early recognition and precise differential diagnosis are critical for guiding appropriate treatment decisions and improving outcomes [13].

The development of artificial intelligence (AI) systems for cardiology differential diagnosis holds significant promise for clinical practice [14,15]. In ICU settings, diagnostic models are expected to directly analyze raw clinical data and generate timely predictions to support clinicians in high-acuity, time-sensitive scenarios [16–18]. In primary care and general cardiology settings, AI systems can assist by integrating longitudinal patient data with extensive medical knowledge sources, such as clinical guidelines and reference literature, to perform comprehensive diagnostic reasoning [19,20]. Such decision-support capabilities may help clinicians reduce diagnostic uncertainty and mitigate the risk of misdiagnosis [21].

However, building reliable AI systems for cardiology differential diagnosis presents several key challenges. First, accurate diagnosis requires deep cardiology-specific knowledge, adherence to clinical guidelines, and the ability to perform sophisticated reasoning to distinguish subtle differences among clinically similar conditions [6]. While numerous diagnostic models have been proposed, most are not tailored specifically to cardiology [22–25], and few are designed to explicitly support complex reasoning over nuanced clinical presentations [26,27], resulting in suboptimal performance in the cardiology domain. Second, trustworthy diagnostic systems should provide transparent and interpretable explanations to support clinician trust and accountability [22,28], rather than solely outputting diagnostic predictions [29–31]. However, many existing models operate as "black boxes", which limits their interpretability and hinders adoption in real-world clinical workflows [28]. Together, these challenges highlight a substantial gap between current AI methodologies and the stringent requirements of practical cardiology applications.

Recent advances in agentic AI offer promising opportunities to address these challenges [32,33]. Agentic approaches can orchestrate multiple specialized modules or sub-agents to collaboratively emulate structured diagnostic reasoning processes [34–36]. By leveraging large language models (LLMs) as a central reasoning backbone, such systems benefit from remarkable natural language understanding and reasoning capabilities, often without requiring extensive task-specific training [37,38]. Furthermore, their modular architectures facilitate integration with heterogeneous clinical data, external medical knowledge, e.g., cardiac guidelines, and decision-support tools, rendering them suited for cardiology settings.

In this study, we introduce HeartAgent, an agentic system specifically designed for trustworthy differential diagnosis in cardiology (Fig. 1). Our contributions are threefold. First, HeartAgent integrates curated cardiology knowledge resources, orchestrates multiple specialized sub-agents to perform complex reasoning, produces transparent reasoning trajectories, and provides references for traceability. Second, we conducted extensive evaluations on three real-world datasets, demonstrating that HeartAgent consistently outperformed comparative methods in both diagnostic accuracy and explanatory quality. In particular, HeartAgent achieved performance improvements exceeding 36% and 20% on the MIMIC dataset [39] and a private dataset from the University of Minnesota, respectively. Third, in expert-AI collaboration settings, clinicians assisted by HeartAgent surpassed independent expert performance by 26.9% in diagnostic accuracy and 22.7% in explanatory quality. Collectively, these results indicate that HeartAgent advances the state of diagnostic AI by delivering a transparent, reliable, and clinically actionable system, thereby supporting trustworthy decision-making in cardiac disease care.

# Results

## Evaluation datasets

We evaluated model performance using multimodal datasets from the Medical Information Mart for Intensive Care (MIMIC) database [39,40], the University of Minnesota clinical data repository (UMN), and the New England Journal of Medicine (NEJM) [41,42]. MIMIC provided clinical notes, ECG waveforms, and echocardiography data [39], while the UMN dataset contributed clinical notes. To evaluate commercial LLMs (e.g., GPT-5 [43]), which are unable to process sensitive clinical records, we constructed a separate dataset from NEJM's publicly available "Case Challenge" series, comprising de-identified clinical narratives and structured laboratory results. Dataset statistics and composition are summarized in Table 1 and Supplementary Note 1. Domain experts annotated the top-3 differential diagnoses and corresponding rationales as ground-truth (Supplementary Note 2).

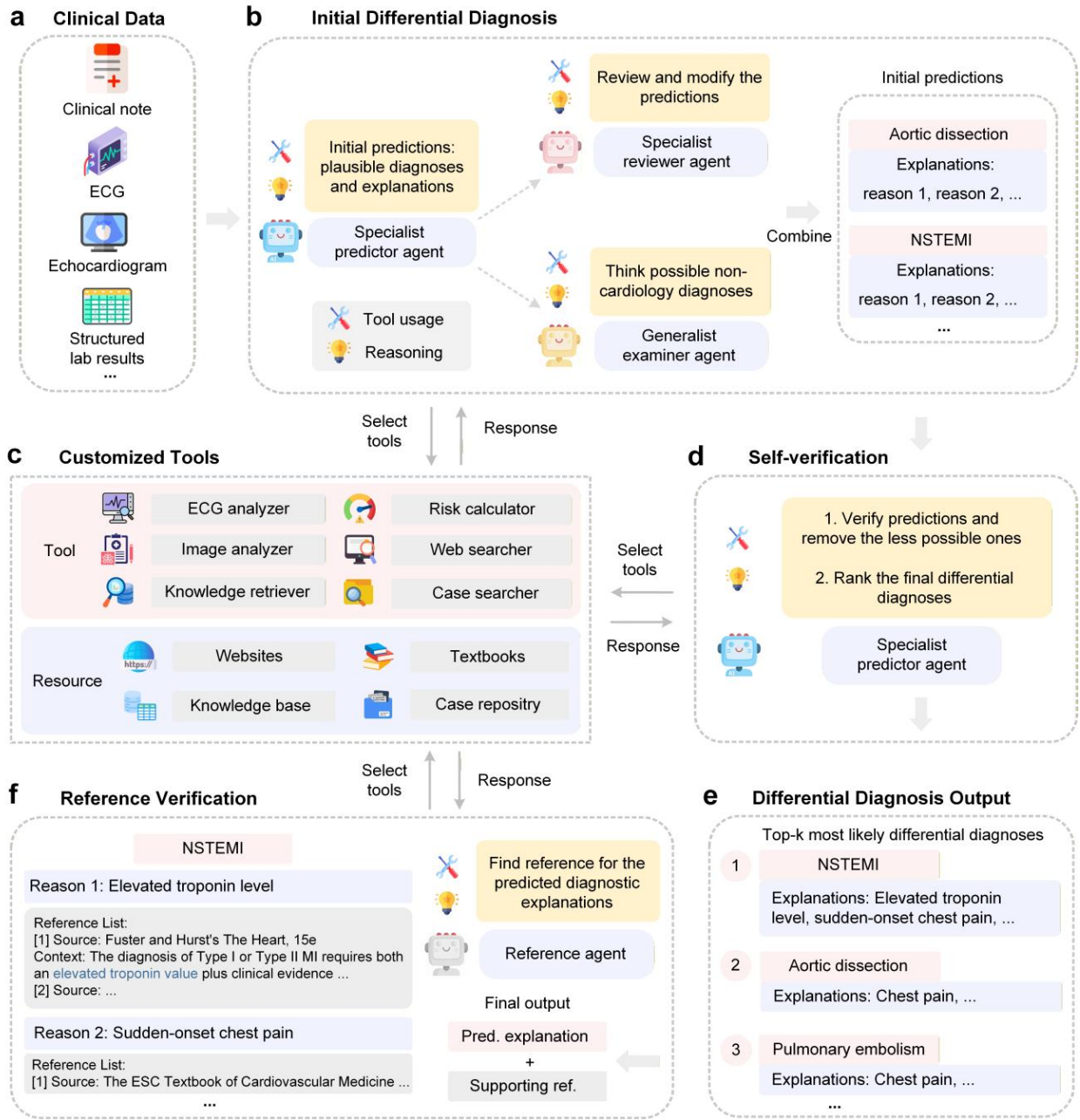

**Fig. 1 | Overview of the HeartAgent framework.** a, Clinical data for cardiac disease diagnosis. b, Initial differential diagnosis stage, which comprises three cooperative agents: a specialist predictor agent that generates initial cardiac diagnostic hypotheses; a generalist examiner agent that incorporates potential

differential diagnoses from non-cardiology specialties; and a specialist reviewer agent that evaluates and refines the initial predictions by supplementing additional plausible cardiac diagnoses with corresponding clinical rationales. c, Customized tools and data resources for cardiology, which can be dynamically invoked during inference to support reasoning and evidence retrieval. d, Self-verification, in which the specialist predictor agent performs iterative self-assessment and refinement of diagnostic hypotheses. e, Differential diagnosis output, presented as a ranked list of the top-k most plausible diagnoses. f, Reference verification, where a reference agent retrieves and validates supporting literature for the generated diagnostic rationales. ECG, electrocardiogram; NSTEMI, non-ST-segment elevation myocardial infarction.

Table 1. Data statistics for the cardiology datasets used in differential diagnosis.

| Statistics | MIMIC | UMN | NEJM |
|---|---|---|---|
| Age (y), mean ± s.d. | 66.0 ± 16.3 | 64.2 ± 24.3 | 53.0 ± 18.8 |
| Male (%) | 50.5% | 52.3% | 73.3% |
| Race (White, Black, Asian, Others), n | (114, 46, 4, 36) | (244, 25, 13, 5)^ | NA |
| Number of cardiology disease types | 56 | 86 | 28 |
| Number of data modalities | 3 | 1 | 2 |
| Sample size | 200 | 300 | 30 |
| Mean note length (words) | 1173.5 | 1220.6 | 618.4 |
| Standard deviation of note length (words) | 499.8 | 242.9 | 203.3 |
| Number of differential diagnoses per note | 3 | 3 | 3 |
| Mean number of explanations per diagnosis | 3.6 | 4.2 | 6.3 |
| Standard deviation of explanations per diagnosis | 0.8 | 1.1 | 2.2 |

*Note: NA, not available. The symbol ^ indicates that data were not available for some subjects.*

## Differential diagnosis performance

We evaluated diagnostic performance across three datasets using multiple base LLMs, including Llama-3.3-70B [44], Qwen-2.5-32B [45], and MedGemma-27B [46]. Top-3 accuracy is reported in Fig. 2a, with top-1 accuracy shown in Supplementary Fig. 1. On the MIMIC dataset, chain-of-thought (CoT) [47] baseline achieved top-3 accuracies ranging from 0.377 (95% confidence interval (CI): 0.332–0.417) for MedGemma-27B to 0.421 (95% CI: 0.435–0.511) for Llama-3.3-70B, whereas HeartAgent improved performance by over 36% across all models ($p < 0.001$). On the NEJM dataset, HeartAgent achieved top-3 accuracies of 0.556 (95% CI: 0.478–0.645) with Llama-3.3-70B, 0.556 (95% CI: 0.467–0.644) with Qwen-2.5-32B, and 0.567 (95% CI: 0.478–0.656) with MedGemma-27B, representing an average improvement of 29% over the strongest baseline. Similarly, on the UMN dataset, HeartAgent achieved top-3 accuracies ranging from 0.560 (95% CI: 0.529–0.590) to 0.592 (95% CI: 0.562–0.622), exceeding baseline methods by more than 20% on average ($p < 0.001$). We further examined the distribution of correct diagnoses within the top-3 predictions. As shown in Fig. 2b for MIMIC (and Supplementary Fig. 2 for other datasets), HeartAgent increased the proportion of cases with three and two correct diagnoses by an average of 9% and 12%, respectively, compared with CoT.

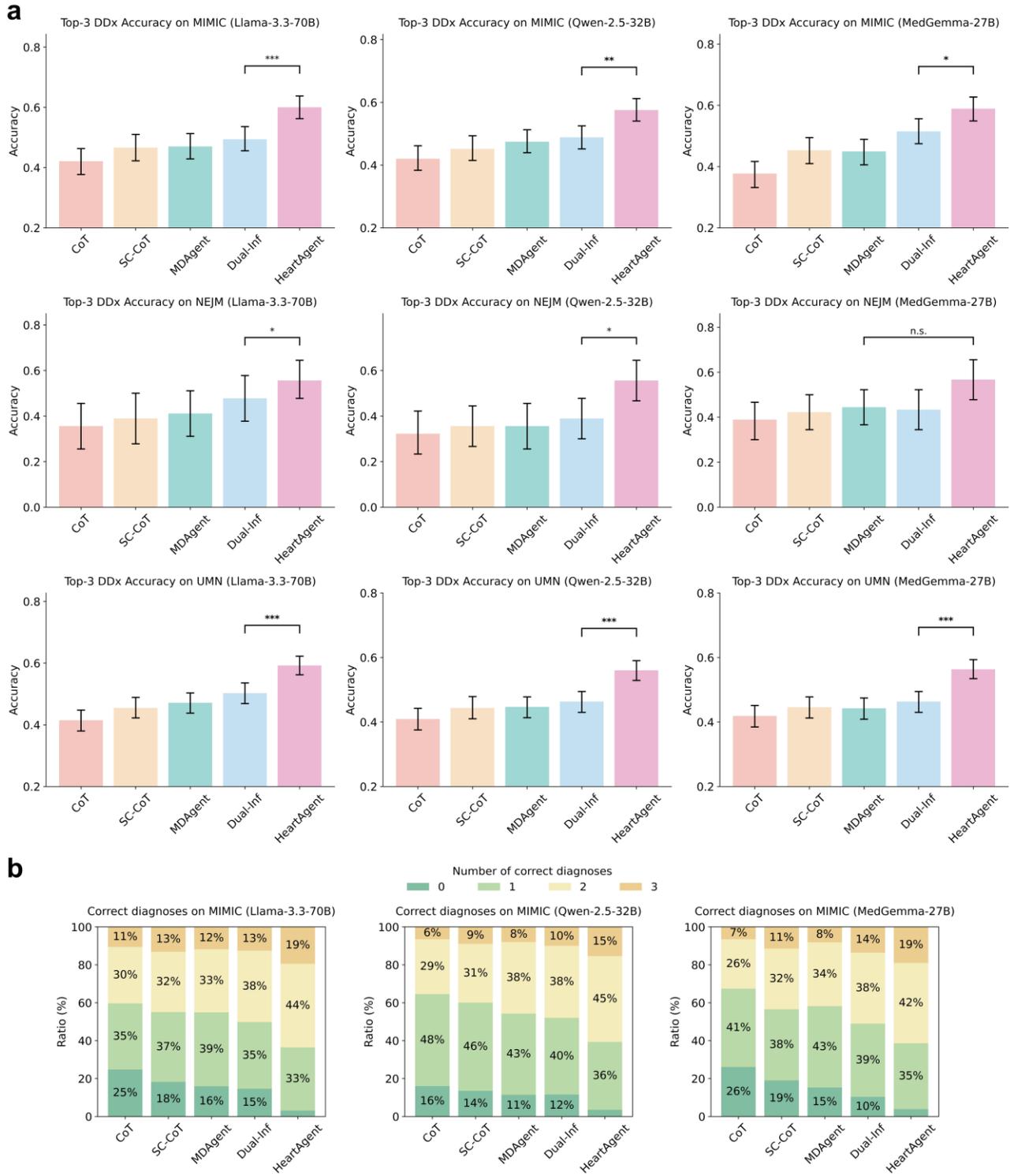

**Fig. 2 | Differential diagnosis performance comparison.** a, Top-3 diagnostic accuracy across the three datasets using different LLMs. "***" denotes $p < 0.001$, "**" denotes $p < 0.01$, "*" denotes $p < 0.05$, and "n.s." denotes not significant ($p > 0.05$), as assessed by a two-sided Mann-Whitney U test. For the NEJM dataset, confidence intervals are reported for consistency but may be less stable due to the limited sample size. b, Distribution of correctly identified diagnoses within the top-3 predictions on the MIMIC dataset.

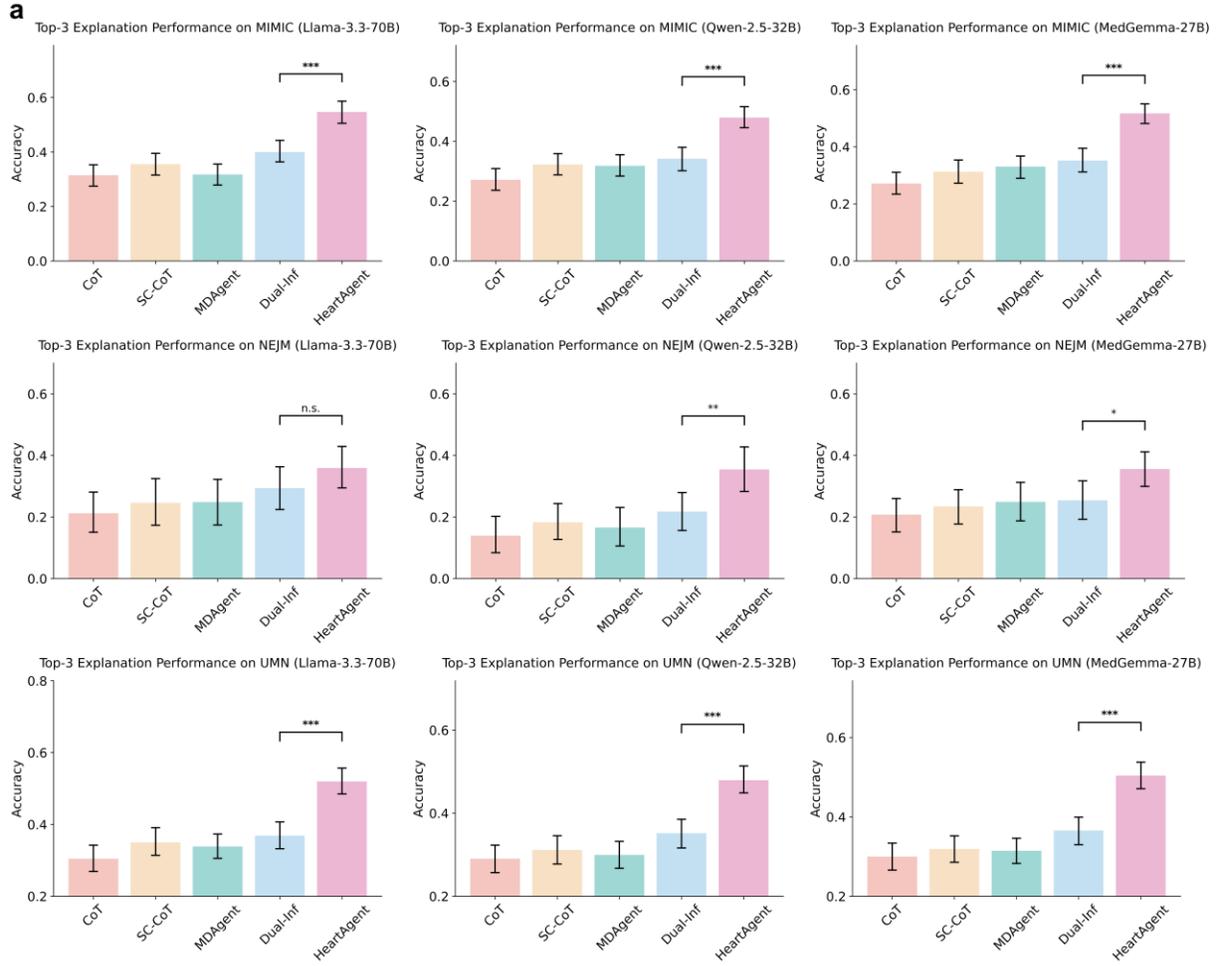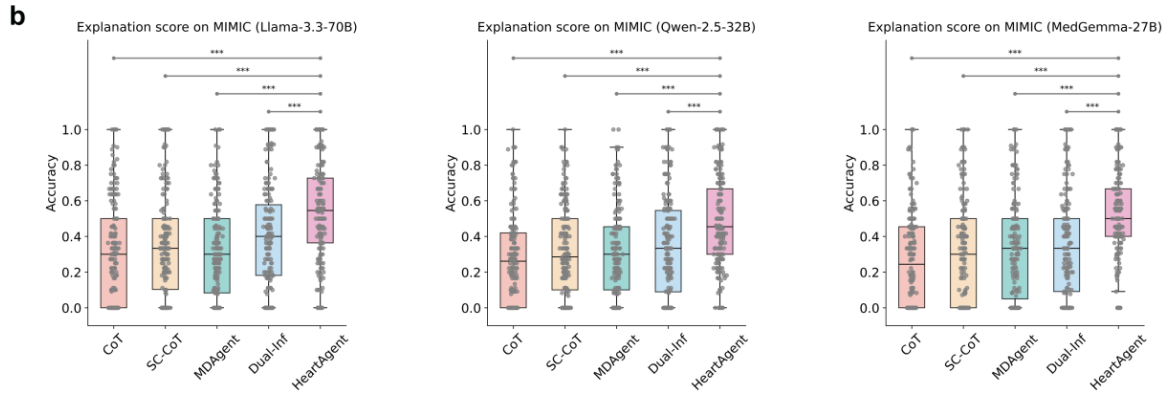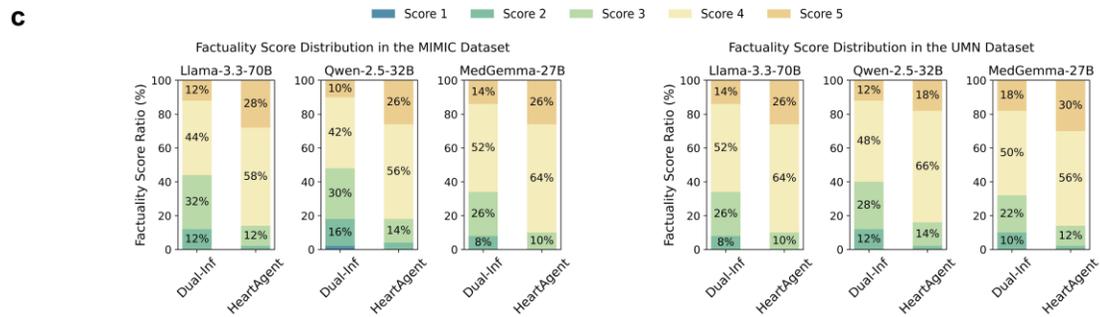

**Fig. 3 | Comparison of diagnostic explanation performance.** a, Explanation performance of top-3 predictions across the three datasets using different base LLMs. Statistical significance is indicated as follows: *** $p < 0.001$, ** $p < 0.01$, * $p < 0.05$, and n.s., not significant ($p > 0.05$), determined using a two-sided Mann–Whitney U test. For the NEJM dataset, confidence intervals are reported for consistency but may be less stable due to the limited sample size. b, Distribution of explanation scores for the top-3 predictions on the MIMIC dataset. c, Comparison of explanation factuality on the MIMIC and UMN datasets. Factuality was assessed using a 5-point Likert scale (1–5), with higher scores indicating greater factual accuracy.

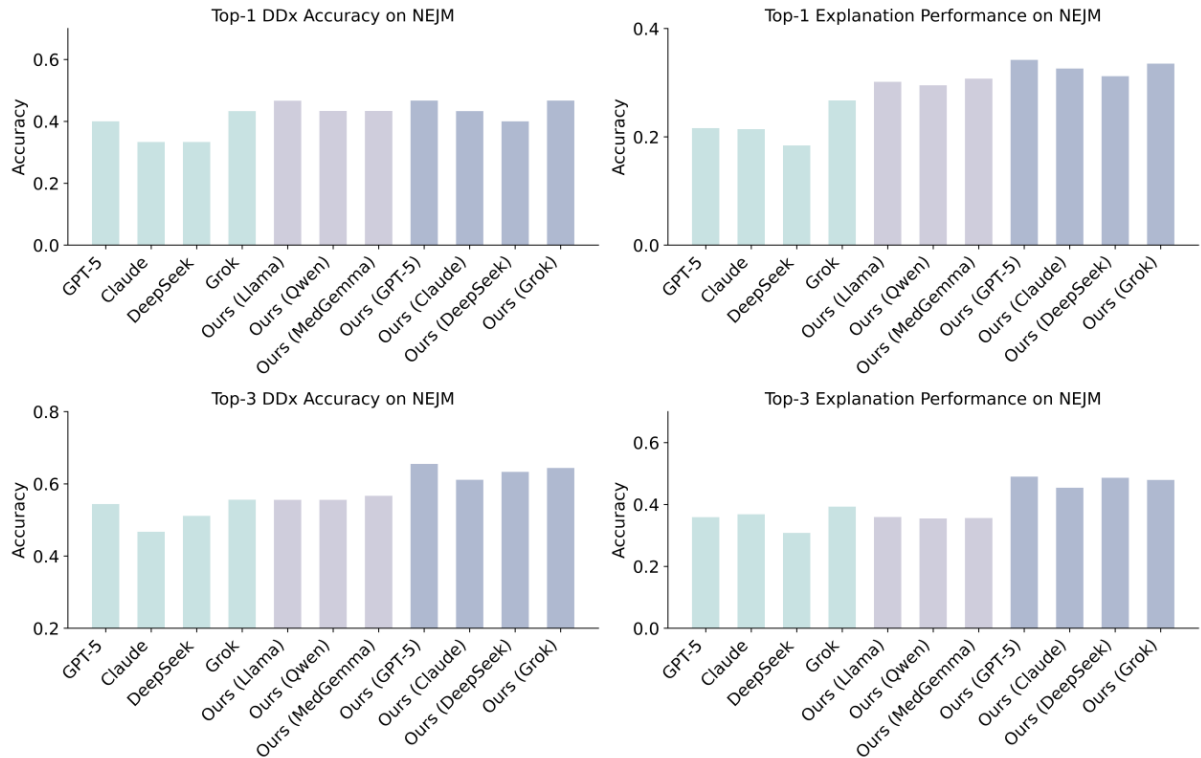

**Fig. 4 | Comparison of differential diagnosis performance between HeartAgent and commercial large language models on the NEJM dataset.** Four commercial LLMs, GPT-5, Claude, DeepSeek-R1, and Grok, each with hundreds of billions of parameters, were evaluated as comparative prediction models. HeartAgent was instantiated using both open-source LLMs with at most 70 billion parameters (Llama-3.3-70B, Qwen-2.5-32B, and MedGemma-27B) and the same commercial LLMs. Across all settings, HeartAgent consistently outperformed the corresponding off-the-shelf models (shown in green) in both differential diagnosis accuracy and explanatory quality.

## Diagnostic explanation performance

We evaluated diagnostic explanation quality across three datasets, reporting top-3 explanation scores in Fig. 3a and top-1 results in Supplementary Fig. 3. On the MIMIC dataset, HeartAgent achieved scores of 0.547 (95% CI: 0.505–0.587) with Llama-3.3-70B, 0.479 (95% CI: 0.445–0.516) with Qwen-2.5-32B, and 0.517 (95% CI: 0.482–0.550) with MedGemma-27B, outperforming comparative methods by 37%,

40%, and 46%, respectively. On the UMN dataset, HeartAgent achieved explanation scores ranging from 0.479 (95% CI: 0.449–0.514) to 0.519 (95% CI: 0.485–0.557), representing an average improvement of 38% over the strongest baseline. Score distributions (Fig. 3b and Supplementary Fig. 4) further demonstrated significantly higher explanation quality for HeartAgent on the MIMIC dataset compared with the runner-up ($p < 0.001$). In addition, qualitative assessment of explanation factuality using a 5-point Likert scale showed that 86% of HeartAgent-generated explanations achieved scores exceeding 4, compared with 56% for Dual-Inf [22] when using Llama-3.3-70B (Fig. 3c). Consistent trends were observed on the UMN dataset, where HeartAgent exceeded Dual-Inf [22] by 24%, 24%, and 18% across the three base LLMs.

## Comparison with commercial LLMs

We evaluated the diagnostic performance of commercial LLMs, comprising hundreds of billions of parameters, on the NEJM dataset. As shown in Fig. 4, standalone commercial LLMs achieved top-1 diagnostic accuracies ranging from 0.333 to 0.433 and top-3 accuracies from 0.467 to 0.556, which were comparable to HeartAgent instantiated with open-source LLMs. Notably, when integrated into the HeartAgent framework, commercial LLMs demonstrated substantial performance gains over their corresponding base models, with average improvements of 18.6% and 22.7% in top-1 and top-3 diagnostic accuracy, respectively. In terms of explanation quality, HeartAgent built on commercial LLMs further outperformed the base models, achieving an average improvement of 34.9% for top-3 explanation performance.

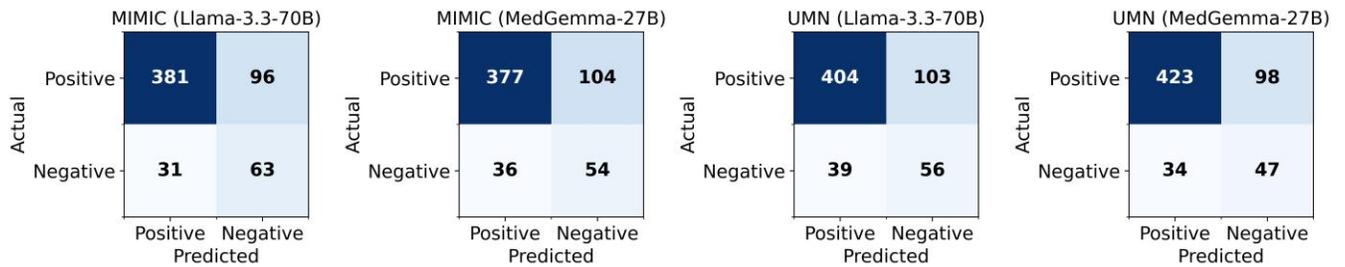

**Fig. 5 | Confusion matrices of reference verification results for HeartAgent.** a–b, Predicted explanations generated by HeartAgent implemented with Llama-3.3-70B and MedGemma-27B on the MIMIC dataset. c–d, Reference verification results on the UMN dataset. A positive label indicates that the retrieved references support a medical statement, while a negative label suggests the absence of supporting evidence.

Table 2. Reference verification performance of HeartAgent on the MIMIC and UMN datasets.

| Dataset (base LLM) | Precision | Recall | $F_1$ |
|---|---|---|---|
| MIMIC (Llama-3.3-70B) | 0.925 | 0.799 | 0.857 |
| MIMIC (MedGemma-27B) | 0.913 | 0.784 | 0.843 |
| UMN (Llama-3.3-70B) | 0.912 | 0.797 | 0.851 |
| UMN (MedGemma-27B) | 0.926 | 0.812 | 0.865 |

**Clinical Text**: A 66-year-old woman was admitted to the hospital because of the nephrotic syndrome. For 30 years, she had had hypertension related to fibromuscular dysplasia of the right renal artery. A saphenous-vein bypass graft had failed, resulting in a poorly functioning right kidney. Angina pectoris developed 24 years before admission. Diabetes mellitus was discovered seven years later and was successfully managed ...
**Structured Lab Results**: ...

**Ground-truth Differential Diagnosis**: Systemic Amyloidosis, Restrictive Cardiomyopathy, Diabetic Nephropathy
**Ground-truth Explanation**: Systemic Amyloidosis: "proteinuria (++++) was present", "nephrotic syndrome" ...

**Specialist Agent:**
**Restrictive Cardiomyopathy.** Explanation: "a cardiac ultrasonographic examination performed at that time showed ...", "very low voltage in the extremity leads", "slight cardiac enlargement", "worsening dyspnea" ...
**Ischemic Cardiomyopathy.** Explanation: "angina pectoris developed 24 years before admission", "a pattern indicative of an old anterior myocardial infarction" ...
**Systemic Amyloidosis.** Explanation: "nephrotic syndrome", "bilateral carpal tunnel syndrome developed", "a 24-hour specimen of urine contained 9570 mg of protein", "nocturnal 'burning pain' in her feet, with loss of vibratory sensation" ...
**Hypertensive Heart Disease.** Explanation: "slight cardiac enlargement", "a new grade 2 systolic murmur was heard", "grade 3 systolic murmur over the precordium and at the apex"

**Specialist Reviewer Agent:**
**Restrictive Cardiomyopathy** (REVISE). Add explanation: "bilateral peripheral edema developed shortly thereafter", "+++ pedal edema", "moderate increase in bilateral pleural effusions"
**Ischemic Cardiomyopathy** (REVISE). Add explanation: "nonspecific ST-segment and T-wave abnormalities", "worsening dyspnea and more frequent anginal attacks"
**Systemic Amyloidosis** (REVISE). Add explanation: "proteinuria (++++) was present, with increasing peripheral edema", "orthostatic hypotension and syncope", "complete atrioventricular block", "slight cardiac enlargement" ...
**Hypertensive Heart Disease** (DELETE). Rationale: The patient lacks the typical structural and electrical features of hypertensive heart disease.
**Secondary Heart Failure** (ADD). Explanation: "bilateral peripheral edema developed shortly thereafter", "+++ pedal edema", "moderate increase in bilateral pleural effusions", "worsening dyspnea", "slight cardiac enlargement"
**Severe Aortic Stenosis** (ADD). Explanation: "a new grade 2 systolic murmur was heard", "angina pectoris developed 24 years before admission", "her blood pressure in the upright position was as low as 60/40 mm Hg" ...
**Constrictive Pericarditis** (ADD). Explanation: "bilateral peripheral edema developed shortly thereafter", "one month later, proteinuria (++++) was present, with increasing peripheral edema" ...

**Generalist Examiner Agent:**
**Diabetic Nephropathy** (ADD). Explanation: "proteinuria (++++) was present, with increasing peripheral edema", "diabetes mellitus was discovered seven years later ...", "a 24-hour specimen of urine contained 9570 mg of protein" ...
**Diabetic Peripheral Polyneuropathy** (ADD). Explanation: "diabetes mellitus was discovered seven years later ...", "nocturnal 'burning pain' in her feet", "vibratory sensation was impaired ...", "loss of vibratory sensation" ...

**Self-verification Stage:**
**Deleted Diagnosis**: Hypertensive Heart Disease, Severe Aortic Stenosis, Constrictive Pericarditis
**Ranked Differential Diagnosis**: Systemic Amyloidosis, Restrictive Cardiomyopathy, Diabetic Nephropathy, Diabetic Peripheral Polyneuropathy, Ischemic Cardiomyopathy, Secondary Heart Failure

**Output:**
**Differential Diagnosis Ranking List**: Systemic Amyloidosis, Restrictive Cardiomyopathy, Diabetic Nephropathy, Diabetic Peripheral Polyneuropathy, Ischemic Cardiomyopathy, Secondary Heart Failure
**Explanation**: Systemic Amyloidosis: ["nephrotic syndrome", "bilateral carpal tunnel syndrome developed" ...] ...

**Reference Verification:**
**Predicted Diagnosis**: Systemic Amyloidosis
**Predicted Explanation 1**: "nephrotic syndrome"
**Reference List: [1] Source**: Fuster and Hursts The Heart. **Context**: Immunoglobulin light-chain amyloidosis ... Presentation can be nonspecific and varied and in addition to HFpEF can include nephrotic range proteinuria, carpal tunnel syndrome, periorbital purpura, macroglossia, diarrhea, and autonomic dysfunction ... **[2] Source**: ...
**Predicted Explanation 2**: "bilateral carpal tunnel syndrome developed"
**Reference List: [1] Source**: Manual of Cardiovascular Medicine, 6e. **Context**: This covers the common clinical presentation of the most common subtypes of systemic amyloidosis, but there is significant overlap in presenting symptoms. Symptoms are also dependent on variant phenotype ... C. Extracardiac symptoms of ATTRwt (can precede cardiac involvement by 5-7 years) 1. Bilateral carpal tunnel syndrome (deposits of amyloid in flexor retinaculum and tenosynovial tissue within the carpal tunnel) ... **[2] Source**: ...
**Predicted Explanation 3**: "a 24-hour specimen of urine contained 9570 mg of protein"
**Reference List**: Not found
...
**Predicted Diagnosis**: Diabetic Nephropathy
**Predicted Explanation 1**: "proteinuria (++++) was present, with increasing peripheral edema"
**Reference List: [1] Source**: The ESC Textbook of Cardiovascular Medicine. **Context**: ACEIs prevent the progression of microalbuminuria to overt proteinuria, and slow the progression of diabetic nephropathy ... **[2] Source**: ...
...

**Fig. 6 | Case studies of HeartAgent on the NEJM dataset.** The base LLM is Llama-3.3-70B. Suggested revisions from the specialist reviewer agent and the generalist examiner agent are shown in purple, correct diagnoses in the final predictions are highlighted in red, and supporting references are highlighted in green. In the self-verification stage, HeartAgent integrated the predictions and suggested revisions, including proposed additions, revisions, and deletions, from all the agents and systematically examined each one. In the reference verification stage, HeartAgent retrieved relevant contexts from medical textbooks or clinical guidelines to support each predicted explanation.

## Reference verification Performance

We evaluated the correctness of references retrieved by HeartAgent through a manual review of 40 cases sampled from the MIMIC and UMN datasets, which encompassed both correct and incorrect predictions. The confusion matrices for Llama-3.3-70B and MedGemma-27B are presented in Fig. 5. Specifically, among the 571 predicted explanations in the sampled data, HeartAgent successfully retrieved supporting references for 412 explanations, of which 381 were verified as correct (Fig. 5a). Meanwhile, HeartAgent failed to retrieve references for 159 predicted explanations; notably, 63 of these explanations were incorrect and therefore had no supporting evidence (Fig. 5a). Furthermore, HeartAgent correctly identified an average of 61% of true negatives, demonstrating that it generally refrains from retrieving references for non-factual explanations. Overall, across both datasets, HeartAgent achieved an average precision of 92% and a recall of 80%, demonstrating strong performance in reference verification (Table 2).

## Case study

Case studies were conducted to demonstrate the effectiveness of HeartAgent in assisting cardiology diagnosis. As illustrated in Fig. 6, the specialist predictor agent generated the initial diagnoses, which covered part of the ground-truth conditions (e.g., restrictive cardiomyopathy). The other two agents further supplemented plausible diagnoses (e.g., diabetic nephropathy) and synthesized diagnostic explanations. In the self-verification stage, HeartAgent integrated the predictions and suggested revisions, including proposed additions, revisions, and deletions, of all agents and systematically examined each one. Through this process, unlikely diagnoses and explanations (e.g., hypertensive heart disease) were filtered out. Ultimately, it correctly identified the differential diagnoses (highlighted in red). In addition, HeartAgent retrieved supporting evidence from medical textbooks or clinical guidelines to substantiate each predicted explanation (highlighted in green). More examples are provided in Supplementary Figs. 5–7.

## AI-augmented clinician assessments

We evaluated HeartAgent's potential to enhance clinical decision-making by comparing its performance with that of human experts. A total of 100 clinical notes from the MIMIC dataset were randomly selected for manual assessment. We compared differential diagnosis performance under three conditions: HeartAgent alone, clinicians alone, and clinicians assisted by HeartAgent, using the Llama-3.3-70B as the base model. As shown in Fig. 7, clinicians outperformed HeartAgent in top-3 diagnostic accuracy (0.650 versus 0.597), while both achieved similar diagnostic explanation quality (0.554 versus 0.534). Notably, HeartAgent-assisted clinicians demonstrated the highest overall performance, with absolute improvements of 21.7% and 26.9% in top-1 accuracy and explanation quality, and 12.3% and 22.7% in top-3 accuracy and explanation quality, respectively, compared with unaided clinicians.

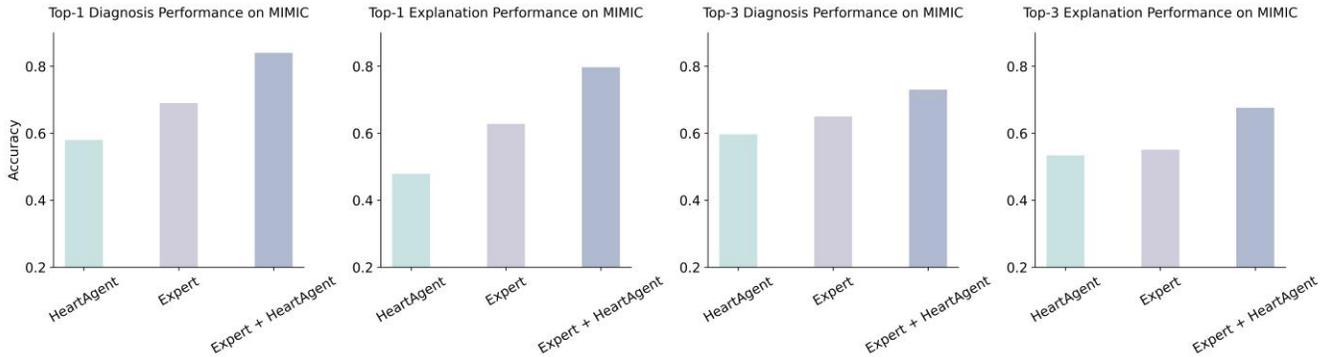

**Fig. 7 | Performance of human–AI collaboration for cardiology differential diagnosis and explanation.** One hundred clinical notes from the MIMIC dataset were randomly selected. HeartAgent was built on Llama-3.3-70B, and the clinician team of two cardiologists provided consensus predictions.

## Ablation study

To assess the contribution of each component in HeartAgent, we performed an ablation study by systematically removing individual agents and resources and evaluating their impact on performance. We examined the roles of the generalist examiner agent and the specialist reviewer agent, as well as the case repository, curated knowledge base, and web resources (Table 3 and Supplementary Table 5). On the MIMIC dataset, removing the generalist examiner agent and specialist reviewer agent led to average declines of 9.5% and 4.9% in top-3 diagnostic accuracy, and 12.3% and 11.5% in explanation performance, respectively. Similarly, restricting access to the case repository, websites, or curated knowledge base resulted in average decreases of 4.6%, 3.6%, and 9.5% in diagnostic performance, and 5.8%, 8.9%, and 18.4% in explanation performance.

Table 3. Top-3 diagnostic accuracy (%) of different HeartAgent variants on the MIMIC dataset.

| Model | Llama-3.3-70B | Qwen-2.5-32B | MedGemma-27B |
| --- | --- | --- | --- |
| HeartAgent | 60.0 [56.2, 63.8] | 57.5 [54.1, 61.2] | 58.9 [54.9, 62.7] |
| w/o Case repository | 57.4 [53.7, 61.2] * | 55.4 [51.9, 59.0] | 55.4 [51.1, 59.2] |
| w/o Corpora | 54.5 [50.8, 58.5] ** | 51.4 [47.9, 55.1] * | 53.8 [49.8, 58.0] |
| w/o Web | 57.6 [53.8, 61.4] * | 55.7 [52.2, 59.4] | 56.7 [52.5, 60.5] |
| w/o GE | 53.0 [49.2, 56.8] *** | 51.7 [48.4, 55.2] * | 54.9 [50.9, 58.9] |
| w/o SR | 56.9 [53.2, 60.5] * | 54.6 [51.1, 58.2] | 56.3 [52.5, 60.3] |

Statistical significance versus HeartAgent was determined using a two-sided Mann-Whitney U test and is indicated by asterisks (* $p < 0.05$, ** $p < 0.01$, and *** $p < 0.001$; Corpora: the collected guidelines and the curated knowledge base; GE: generalist examiner agent; SR: specialist reviewer agent).

# Discussion

First, HeartAgent demonstrated superior diagnostic performance compared with baseline methods (Fig. 2). As shown in Fig. 2b, it produced more accurate differential diagnoses, achieving higher top-1 and top-3 accuracy across the three datasets. This improvement stems from two main factors. First, HeartAgent can dynamically access external knowledge from a curated, domain-specific knowledge base or websites, rather than relying solely on the parametric knowledge of LLMs, bridging gaps in cardiology expertise. Second, the framework is specifically designed for differential diagnosis, with collaborative agents fulfilling distinct roles. In particular, the generalist examiner agent and specialist reviewer agent broaden the scope of potential diagnoses, enhancing overall diagnostic performance. These capabilities suggest that HeartAgent, especially its high top-3 accuracy, can serve as a valuable reference for cardiologists during differential diagnosis.

Second, HeartAgent excelled in explanatory capabilities. Across the top-3 predictions, it achieved average performance gains of over 41% and 38% relative to baseline methods on the MIMIC and UMN datasets, respectively (Fig. 3a). Its explanations also demonstrated higher factuality compared with a strong comparative method (Fig. 3c), indicating a reduced risk of hallucinations. These advantages arise from two factors. First, HeartAgent retrieves accurate cardiology knowledge from curated corpora or websites to support diagnostic reasoning. Second, its agent-based workflow, where the three agents examine, refine, and supplement explanations, ensures higher quality output. The robust explanatory performance can help clinicians assess the trustworthiness of predictions from otherwise black-box models [28].

Third, HeartAgent exhibited strong generalizability across base LLMs. Whether using MedGemma-27B, Llama-3.3-70B, or commercial LLMs such as DeepSeek-R1 [48], it consistently outperformed baseline models in both diagnosis and explanation (Figs. 2–4). Notably, HeartAgent, built on smaller models, such as MedGemma-27B, achieved performance comparable to or exceeding that of large-scale models with hundreds of billions of parameters, including GPT-5 [43] and DeepSeek-R1 [48] (Fig. 4). With large-scale LLMs, HeartAgent further surpassed all commercial models (Fig. 4). This performance can be attributed to two primary factors: commercial LLMs provide strong reasoning and planning capabilities, and large-scale models possess extensive medical knowledge, offering a robust foundation for HeartAgent to achieve high accuracy.

HeartAgent also provided reliable verification of the generated diagnostic explanations. Among the identified references, 92% of them are correct and used to support the predicted explanation (Fig. 5). This reliability is primarily due to the extensive cardiology corpus, including textbooks and guidelines, used as reference sources. The significance of the reference verification is two-fold: it confirms the correctness of explanations, thereby supporting accurate differential diagnoses, and it provides evidence to assist clinicians in decision-making and validation of model outputs. However, HeartAgent failed to retrieve supporting references for around 20% of explanations in the sampled data. This is primarily because some clinically correct explanations are expressed in non-canonical forms that differ from textbook terminology. For example, "estimated valve area was 0.6 cm²" corresponds to "valvular stenosis" in guidelines, but the lexical and semantic mismatch via BM25 [49] and MedCPT [50] hampers direct context retrieval. This observation suggests that additional normalization or transformation of model-generated explanations into canonical expressions may improve evidence retrieval.

Additionally, HeartAgent enhanced clinicians' performance in differential diagnosis and explanation (Fig. 7). By providing a set of plausible diagnoses, corresponding explanations, and reference verification, HeartAgent reduces the risk of missing critical diagnoses or relevant rationales. This support also helps cardiology clinicians identify the most probable diagnosis, explaining the substantial improvement in top-1 accuracy (Fig. 7). Specifically, in one case, although HeartAgent's top-3 predictions encompassed STEMI, Takotsubo syndrome, and acute pericarditis, the clinicians found Takotsubo syndrome was more plausible based on the model's explanations, such as "an identifiable acute stress trigger", and correctly ranked it as the most likely diagnosis. Overall, human–AI collaboration combines the advantages of the extensive knowledge capacity of HeartAgent and the complex reasoning ability of clinicians, resulting in superior accuracy and more reliable explanations.

Several factors influenced HeartAgent's performance. Base LLMs determine the agents' reasoning ability and parametric knowledge capacity, directly impacting overall performance. In our study, Llama-3.3-70B and MedGemma-27B achieved comparable results and outperformed Qwen-2.5-32B in both diagnosis and explanation (Figs. 2–3). Considering hardware constraints in clinical settings, domain-specific LLMs with strong reasoning capability, such as MedGemma-27B, are recommended. The curated corpora, which are used to construct the customized knowledge base, significantly enhance performance by supplying cardiology-specific knowledge (Table 3 and Supplementary Table 5). This is because the off-the-shelf LLMs lack sufficient cardiology knowledge, while the summarized knowledge, such as diagnostic criteria in the knowledge base, provides concise and accurate information, thus facilitating HeartAgent's diagnostic reasoning. Similarly, access to web resources is beneficial (Table 3). This is because, for rare cardiac diseases, such as Kounis syndrome, or non-cardiac differential diagnoses, such as seizure and esophageal spasm, which may not be covered in the customized knowledge base, HeartAgent could search online information to bridge the gap. Additionally, the diversity of the case repository is also conducive, as it provides reference materials for decision-making.

Despite these advances, the study has several limitations. First, all evaluation data were derived from adult patients, leaving the application to pediatric populations unexplored. Second, due to privacy restrictions, performance with commercial LLMs was only evaluated on open-source datasets. Future work could assess commercial LLMs in HIPAA-compliant environments.

In summary, this study addresses the pressing need for trustworthy and explainable models in cardiology differential diagnosis. We propose HeartAgent, an autonomous agent system that coordinates multiple agents, leverages customized tools, and integrates diverse external knowledge to generate accurate differential diagnoses, explanations, and verified references. HeartAgent outperforms comparative methods across three datasets, demonstrates robust performance across various base LLMs, and significantly enhances clinician performance in human–AI collaboration. Overall, this study advances the trustworthiness of AI models in cardiology disease diagnosis, securing reliable and evidence-based clinical decision-making.

# Methods

## Dataset curation and annotation

The MIMIC-IV database [39,40] contains de-identified electronic health records (EHRs) from nearly 300,000 patients treated at Beth Israel Deaconess Medical Center in Boston, Massachusetts, between 2008 and 2019. This publicly available resource includes multiple clinical data types [51], such as narrative notes, ECGs, and echocardiograms, capturing information on laboratory findings, diagnoses, and procedures. The NEJM case reports were sourced from the New England Journal of Medicine "Case Challenge" series [6,41,42], an educational resource presenting detailed clinical scenarios with multiple data types, including narrative notes, CT scans, ECGs, X-rays, and echocardiograms, followed by expert commentary to guide clinical diagnosis. Thirty cases were selected from this series. The UMN database is a private collection of EHRs for approximately 1,200,000 patients treated at the University of Minnesota Medical Center between 2011 and 2024, providing comprehensive clinical information, including chief complaints, laboratory results, and diagnostic records. Further details are provided in Supplementary Note 1.

Although the raw data contain ground-truth diagnoses, detailed annotation was required for this study. Three domain experts manually reviewed each case to curate the top-3 differential diagnoses and the corresponding explanations. The full annotation guidelines are provided in Supplementary Note 2.

## Overview of HeartAgent

We propose HeartAgent, a customized multimodal, multi-agent framework for cardiology differential diagnosis. The system is designed to analyze heterogeneous clinical data, perform self-verified differential diagnosis, and provide reference-supported rationales. HeartAgent consists of four collaborative agents: (1) a specialist predictor agent, which generates initial diagnostic predictions and conducts self-verification and refinement; (2) a generalist examiner agent, which considers differential diagnoses from non-cardiology specialties to reduce misdiagnosis risk; (3) a specialist reviewer agent, which evaluates initial predictions and supplements additional plausible cardiac diagnoses with explanations; and (4) a reference verification agent, which retrieves references to support the diagnostic rationales.

The overall workflow is illustrated in Fig. 1. Multimodal clinical inputs, such as ECGs, echocardiograms, and structured laboratory results, first pass through modality-specific tools, such as an ECG interpreter, echocardiography analyzer, tabular data processor, and risk calculator, to generate standardized reports. In the next stage, the specialist predictor agent integrates these processed data to produce initial diagnostic predictions and explanatory rationales. To broaden the diagnostic scope and reduce potential errors, the generalist examiner agent explores differential diagnoses beyond cardiology, while the specialist reviewer agent reassesses initial predictions and proposes additional relevant cardiac conditions. Both agents leverage external knowledge sources, including structured databases, textbooks, clinical guidelines, and web searches, to retrieve representative clinical presentations for candidate diseases. The third stage, self-verification, involves the specialist predictor agent systematically evaluating each candidate diagnosis, eliminating less plausible options, and refining remaining predictions. This process uses diverse information sources, including case repositories, structured databases, and medical websites, to retain the most likely diagnoses and refine their rationales. Finally, in the reference verification stage, the reference verification agent retrieves supporting evidence for the predicted explanations from curated, domain-specific corpora, enhancing the transparency and reliability of HeartAgent's outputs.

## HeartAgent composition: tools

To extend the functional capacity of the agent-based system, HeartAgent is augmented with a suite of external, modality-specific tools that can be dynamically invoked during inference.

For medical image interpretation, including echocardiography, chest X-rays, CT scans, and ECG images, HeartAgent dispatches the input to the corresponding analyzer implemented with a vision-language model. Each analyzer is prompted with modality-specific instructions to generate structured, comprehensive reports, capturing salient visual findings, clinically relevant measurements (e.g., chamber dimensions or waveform intervals when available), and qualitative uncertainty estimates. When longitudinal or multi-view images are provided, each image is processed independently, and the extracted findings are aggregated to produce a comparative assessment.

For waveform-based ECG analysis, HeartAgent incorporates an ECG waveform interpreter to process raw or digitized signals. Implemented using the NeuroKit2 library [52], this interpreter performs automated signal preprocessing, peak detection, and waveform delineation. It identifies key ECG components, including P waves, QRS complexes, and T waves, and derives beat-level and rhythm-level features, such as heart rate variability, RR interval statistics, and rhythm irregularities. Outputs are provided as structured JSON objects containing quantitative features alongside concise narrative summaries describing electrophysiological characteristics.

To facilitate early risk stratification of cardiac diseases, HeartAgent further includes a cardiology-specific risk calculator. This module automatically extracts relevant demographic and clinical variables from unstructured patient text using an LLM-based parser, followed by rule-based validation. Risk scores are computed according to expert-curated clinical rubrics, reflecting the estimated likelihood of developing specific cardiac conditions at an early stage. In total, twenty cardiology-related risk scoring schemes were curated from MDCalc [53] (Supplementary Note 4).

To address potential knowledge gaps, HeartAgent is equipped with a knowledge retriever that accesses cardiology-specific diagnostic information from structured databases, textbooks, and professional guidelines. Queries to structured databases retrieve diagnostic criteria, representative symptoms, common differential diagnoses, and distinguishing features by matching disease entities. To overcome limitations such as synonym variation and hierarchical taxonomy, an LLM-based entity normalization step maps queried diseases to semantically related concepts in the knowledge base.

When information from the structured knowledge base is insufficient, the system activates a web searcher to retrieve supplementary medical knowledge from online sources. The web searcher constructs tailored query strings combining candidate disease names with agent-generated keywords and submits them to the Wikipedia and PubMed APIs. In practice, only the top three documents from each source are retained. Documents are segmented into text chunks and summarized using an LLM to extract diagnostic knowledge, including disease definitions, diagnostic criteria, and representative clinical manifestations.

To leverage empirical clinical experience, HeartAgent incorporates a case searcher that retrieves similar patient cases from a curated cardiology-specific case repository. All repository cases are preprocessed and encoded offline into fixed-dimensional vector embeddings using a pre-trained language model to enable efficient similarity search. During inference, the input patient note undergoes the same preprocessing and embedding procedure. Cosine similarity scores are then computed between the query embedding and stored case embeddings, and the top five most similar cases are retrieved. Each case is returned with its confirmed diagnosis and a concise clinical summary generated by an LLM, allowing the diagnostic agents to reference analogous real-world cases and incorporate empirical experience.

## HeartAgent composition: data resources

To support cardiology differential diagnosis, we constructed a set of customized data resources, including cardiology-specific corpora (e.g., guidelines and textbooks), structured knowledge databases, and a case repository.

We manually collected 109 cardiology-related medical guidelines from professional societies and authoritative clinical organizations, along with 10 standard cardiology textbooks (Supplementary Note 9). All PDF documents were converted to plain text. These resources served two purposes: building the structured knowledge database and providing standard references for the reference verification stage.

The structured knowledge database is used to provide extensive cardiology-specific diagnostic knowledge to assist the agents in decision-making (Supplementary Note 10). Experts systematically reviewed the collected corpus and populated the knowledge base with diagnosis-oriented knowledge, including representative clinical presentations, formal diagnostic criteria, common differential diagnoses, and distinguishing features for differentiating competing conditions. Knowledge entries were manually extracted with consensus checking to ensure internal consistency and clinical validity. The knowledge base encompasses over 107 cardiac diseases and 41 non-cardiac differential diagnoses with overlapping or confounding presentations. All entries are stored in JSON format to enable structured querying by disease name.

We also constructed a cardiology-specific case repository containing real-world clinical cases to provide empirical references. The repository covers 110 disease types and includes 4000 clinical notes derived from the MIMIC-IV database. To prevent data leakage, patient-level separation was enforced between the case repository and the evaluation datasets, ensuring no patient contributed notes to both sets. During preprocessing, clinical notes were segmented into predefined sections using rule-based and keyword-based heuristics. Sections describing diagnostic findings and multimodal report analyses (e.g., CT, echocardiography, and ECG interpretations) were retained, whereas treatment, hospital course, and discharge summaries were removed to reduce noise. Retained cases were standardized into a unified textual format and indexed for embedding-based similarity search by the case searcher.

## HeartAgent composition: reference verification

The reference agent enhances diagnostic explanations with explicit citations by cross-referencing a curated corpus of medical guidelines and textbooks. All source PDFs were converted to plain text and segmented into overlapping chunks of 800 words with a 50-word stride to preserve contextual continuity. Each chunk was indexed with structured metadata, e.g., source title, enabling precise evidence traceability.

Given a candidate diagnosis and its explanation, the agent first employs an LLM to rewrite the explanation as a concise, retrieval-oriented medical claim. To ground this claim in authoritative medical knowledge, we adopt a two-stage retrieval strategy. In the first stage, BM25 [49] retrieves a high-recall set of candidate passages, selecting the top 20 lexically relevant chunks based on the LLM-generated query. In the second stage, MedCPT [50] encodes both the query and BM25-retrieved passages, re-ranking candidates according to dense embedding similarity to refine results based on biomedical semantic relevance.

The top five re-ranked chunks are subsequently passed back to the LLM, which performs two tasks: (1) determining whether a passage contains contextually relevant evidence that supports the explanation-diagnosis relationship, and (2) when such evidence is present, extracting the supporting text along with its associated source metadata, e.g., book title. This hybrid retrieval-and-verification design combines the robustness of lexical matching with the precision of semantic retrieval, enabling reliable and explainable diagnostic prediction.

## Implementation details

HeartAgent was implemented using multiple LLMs to systematically evaluate robustness and generalizability. Due to ethical and data governance restrictions associated with the MIMIC and UMN datasets, experiments on these datasets were conducted exclusively with open-source LLMs. Specifically, Llama-3.3-70B [44], Qwen-2.5-32B [45], and MedGemma-27B [46] were used as backbone models for the agent modules. For the publicly available NEJM case reports, we additionally evaluated proprietary frontier models, including GPT-5 [43], DeepSeek-R1 [48], claude-3-7-sonnet-20250219 (Claude), and grok-4-fast-reasoning (Grok), to assess upper-bound performance under less restrictive conditions. Unless otherwise specified, all agent prompts were executed with a temperature of 0.1 to balance reasoning diversity and output stability. For deterministic components, namely the tabular data analyzer, knowledge searcher, web searcher, and case searcher, the temperature was set to zero. Open-source LLMs used in these tools were deployed with 4-bit quantization to reduce GPU memory consumption while maintaining inference accuracy.

For multimodal image interpretation, including analyzers for echocardiography, ECG, chest X-rays, and CT scans, vision-language inference was implemented using MedGemma-4B [46]. Each modality-specific analyzer was prompted with standardized, task-oriented instructions to generate structured reports capturing salient visual findings and clinically relevant measurements, when available. The ECG waveform interpreter was implemented using the NeuroKit2 Python library [52] with default configurations for signal preprocessing, R-peak detection, and waveform delineation. Derived outputs included beat-level and rhythm-level features, such as heart rate variability indices and RR interval statistics, returned as structured outputs.

For the case searcher, Clinical-Longformer [54] was used to compute document-level embeddings for all clinical notes in the repository. Notes were preprocessed using the same pipeline as at inference time to ensure embedding consistency. Cosine similarity was used as the distance metric, and the top five most similar cases were retrieved for each query.

The reference verification module employed a hybrid retrieval strategy combining BM25 [49] and MedCPT [50]. BM25 [49] was first applied to obtain a high-recall candidate set of passages, which were subsequently re-ranked using dense embeddings generated by MedCPT [50] to enhance biomedical semantic relevance. Default hyperparameters were used for both retrievers.

For comparative evaluation, several baseline diagnostic methods were implemented, including CoT [47], Self-Consistency CoT (SC-CoT) [55], Dual-Inf [22], and MDAgent [26]. Implementations closely followed the configurations and prompting strategies described in the original studies. Specifically, SC-CoT [55] generated five independent reasoning trajectories per clinical note, and final diagnoses and explanations were determined by majority agreement.

## Collaboration of HeartAgent and clinicians

To evaluate the impact of human–AI collaboration on diagnostic decision-making, HeartAgent was integrated into clinicians' review workflows. In this setup, the model's predictions served as an additional source of diagnostic insight. Clinicians could interact iteratively with HeartAgent by issuing instructions, requesting clarifications, and refining diagnostic hypotheses, while the system generated corresponding differential diagnoses, explanations, and supporting evidence. The clinician team consisted of two cardiology physicians who, with access to medical references and professional guidelines, produced consensus predictions. For evaluation, 100 cases were randomly sampled from the MIMIC dataset, with stratification across disease categories to ensure representative coverage.

## Evaluation metrics

HeartAgent was evaluated using diagnostic accuracy, explanation quality, and reference verification. Consistent with prior studies [6,56,57], top-1 and top-3 accuracy were adopted as primary metrics for diagnostic performance. Top-1 accuracy measures whether the highest-ranked predicted diagnosis matches the ground truth, while top-3 accuracy assesses whether the correct diagnosis appears among the three most confident predictions. Two domain experts manually graded the differential diagnosis accuracy.

Explanation performance was assessed using both quantitative and qualitative metrics. Quantitative accuracy was measured by comparing model-generated diagnostic rationales with ground-truth rationales annotated by domain experts, following related studies [20] and using the LLM-as-a-Judge approach [58,59] (Supplementary Note 5). In addition, we assessed factuality to quantify the presence and severity of inaccurate, misleading, or fabricated information in the explanations. Two physicians independently reviewed each explanation and rated its factual accuracy on a five-point Likert scale: 1, severely inaccurate with three or more distinct factual errors; 2, largely inaccurate with two factual errors; 3, minimally accurate with one factual error; 4, largely accurate with no factual errors but minor imprecision that does not affect core correctness; 5, fully accurate with all statements factually correct and evidence-based. Disagreements were resolved by consensus.

The reliability of reference verification was evaluated via manual expert review. Each reference was examined to determine whether it genuinely supported the corresponding diagnostic statement. Precision, recall, and $F_1$ score were computed [60,61], defining true positives as correctly retrieved references that support valid statements, false positives as retrieved references that do not support the statements, false negatives as missing references for correct statements that could be supported by authoritative sources, and true negatives as the absence of references for incorrect statements. Precision measured the proportion of correct references among all retrieved references, recall measured the proportion of supportable statements with correct references retrieved, and the $F_1$ score was calculated as the harmonic mean of precision and recall.

## Data Availability

All data generated during this study are presented in the Supplementary Notes. The MIMIC-IV dataset is hosted on PhysioNet (https://physionet.org/content/mimiciv/3.1), and access is subject to the platform's data use requirements. The UMN dataset was collected from real-world clinical settings and approved by the appropriate Institutional Review Board; however, the underlying clinical notes cannot be shared publicly due to patient privacy protections. Clinical case reports from the New England Journal of Medicine (NEJM) are available at https://www.nejm.org/browse/nejm-article-category/clinical-cases. The aggregated NEJM cases used in this study, together with manual annotations, will be released publicly upon acceptance of the manuscript.

## Author Contributions

S.Z. conceived the project and oversaw the overall study design. S.Z., K.Y., S.W., and Z.Z. performed the literature review, and S.Z., S.W., and W.X. led the development of the model. Data annotation planning and human evaluation were coordinated by S.Z. and R.Z. Raw data acquisition, preprocessing, and statistical analysis were carried out by S.Z., S.H., H.Z., K.Y., M.Z., and Z.Z., while M.T., Y.C., and H.Z. contributed to data annotation and evaluation. The experimental framework was designed by S.Z., K.Y., F.X., L.C., and R.Z. S.Z. prepared the first draft of the manuscript, with B.R., F.X., and R.Z. providing supervision throughout the project. All authors engaged in scientific discussions, reviewed and revised the manuscript, and approved the final version.

## Acknowledgment

This study was funded by the National Center for Complementary and Integrative Health (R01AT009457), the National Institute on Aging (R01AG078154), and the National Cancer Institute (R01CA287413), all within the National Institutes of Health. The views expressed are those of the authors and do not necessarily reflect the official positions of the NIH. We also acknowledge support from the Center for Learning Health System Sciences at the University of Minnesota.

# Competing Interests

The authors declare no competing interests.